\documentclass{article}
\usepackage[utf8]{inputenc}
\usepackage[numbers]{natbib}
\usepackage{xcolor}
\usepackage[normalem]{ulem}
\usepackage[colorlinks]{hyperref}       
\usepackage{url}            
\usepackage{booktabs}       
\usepackage{amsfonts}       
\usepackage{csquotes}
\usepackage{amsmath}
\usepackage[margin=1.25in]{geometry}

\begin{document}

\title{%
Troubling Trends in Machine Learning Scholarship
}
\author{
 Zachary C. Lipton
 \thanks{Equal Authorship}  
\enskip \& Jacob Steinhardt
\footnotemark[1] \\
Carnegie Mellon University
, Stanford University
\\ 
\href{mailto:zlipton@cmu.edu}{\nolinkurl{zlipton@cmu.edu}},
\href{mailto:jsteinhardt@cs@cs.stanford.edu}{\nolinkurl{jsteinhardt@cs.stanford.edu}}
}

\maketitle
\section{Introduction}
\label{sec:introduction}
Collectively, machine learning (ML) researchers 
are engaged in the creation and dissemination of knowledge about data-driven algorithms.
In a given paper, researchers might aspire to any subset of the following goals, among others:
to theoretically characterize what is learnable, 
to obtain understanding through empirically rigorous experiments, 
or to build a working system that
has high predictive accuracy. 
While determining which knowledge 
warrants inquiry may be subjective,
once the topic is fixed,
papers are most valuable to the community 
when they act in service of the reader,
creating foundational knowledge
and communicating as clearly as possible. 

What sort of papers best serve their readers?
We can enumerate desirable characteristics:
these papers should (i) provide intuition 
to aid the reader's understanding, 
but clearly distinguish it 
from stronger conclusions supported by evidence; 
(ii) describe empirical investigations that
consider and rule out alternative hypotheses \citep{platt1964strong};
(iii) make clear the relationship between theoretical analysis and intuitive or empirical claims
\citep{romer2015mathiness}; and
(iv) use language to empower the reader,
choosing terminology to avoid 
misleading or unproven connotations, 
collisions with other definitions, 
or conflation with other related but distinct concepts \citep{minsky2007emotion}.

Recent progress in machine learning comes despite frequent departures from these ideals.
In this paper, we focus on the following four patterns 
that appear to us to be trending  in ML scholarship: 
\begin{enumerate}
\item Failure to distinguish
between explanation and speculation.
\item Failure to identify the sources of empirical gains, 
e.g.~emphasizing unnecessary modifications 
to neural architectures 
when gains actually stem from hyper-parameter tuning.
\item Mathiness: the use of mathematics that obfuscates or impresses rather than clarifies,
e.g.~by confusing technical and non-technical concepts.
\item Misuse of language, e.g.~by choosing terms of art with colloquial connotations 
or by overloading established technical terms.
\end{enumerate}
 
While the causes behind these patterns are uncertain, 
possibilities include the rapid expansion of the community, 
the consequent thinness of the reviewer pool, 
and the often-misaligned incentives between scholarship 
and short-term measures of success (e.g.~bibliometrics, attention, and entrepreneurial opportunity).
While each pattern offers a corresponding remedy 
(don't do it), 
we also discuss some speculative suggestions 
for how the community might combat these trends. 

As the impact of machine learning widens, 
and the audience for research papers 
increasingly includes students, journalists, and policy-makers,
these considerations apply to this wider audience as well.
We hope that by communicating 
more precise information with greater clarity, 
we can accelerate the pace of research,
reduce the on-boarding time for new researchers, 
and play a more constructive role in the public discourse.

Flawed scholarship threatens 
to mislead the public 
and stymie future research 
by compromising ML's intellectual foundations.
Indeed, many of these problems have recurred cyclically throughout the history of artificial intelligence and, more broadly,
in scientific research.
In 1976, Drew McDermott \citep{mcdermott1976artificial} chastised the AI community for abandoning self-discipline,
warning prophetically that ``if we can't criticize ourselves, 
someone else will save us the trouble''. 
Similar discussions recurred throughout 
the 80s, 90s, and aughts 
\citep{cohen1988evaluation, korf1997does, armstrong2009improvements}. 
In other fields such as psychology, 
poor experimental standards
have eroded trust in the discipline's authority \cite{open2015estimating}.
The current strength of machine learning owes to a large body of rigorous research to date, both theoretical \cite{freund1997decision,bottou2008tradeoffs,duchi2011adaptive} and empirical \citep{jarrett2009best, glorot2010understanding,bergstra2012random}. 
By promoting clear scientific thinking and communication, 
we can sustain the trust and investment 
currently enjoyed by our community.

\section{Disclaimers}
\label{sec:disclaimers}
This paper aims to instigate discussion,
answering a call for papers 
from the ICML Machine Learning Debates workshop.
While we stand by the points represented here, 
we do not purport to offer a full or balanced viewpoint 
or to discuss the overall quality of science in ML.
In many aspects, such as reproducibility, 
the community has advanced standards 
far beyond what sufficed a decade ago.
We note that these arguments 
are made by \emph{us}, against \emph{us},
by insiders offering a critical introspective look, 
not as sniping outsiders.
The ills that we identify are not specific 
to any individual or institution.
We ourselves have fallen into these patterns,
and likely will again in the future. 
Exhibiting one of these patterns doesn't make a paper \emph{bad} nor does it indict the paper's authors, however we believe that all papers could be made stronger by avoiding these patterns. 
While we provide concrete examples, 
our guiding principles are
to (i) implicate ourselves,
and (ii) to preferentially select 
from the work of better-established researchers 
and institutions that we admire, 
to avoid singling out junior students 
for whom inclusion in this discussion 
might have consequences and who lack the opportunity to reply symmetrically. 
We are grateful to belong to a community 
that provides sufficient intellectual freedom 
to allow us to express critical perspectives.

\section{Troubling Trends}
\label{sec:troubling_trends}
In each subsection below, we  
(i) describe a trend; 
(ii) provide several examples (as well as positive examples that resist the trend); 
and (iii) explain the consequences. 
Pointing to weaknesses in individual papers can be a sensitive topic. To minimize this, we keep examples short and specific.

\subsection{Explanation vs. Speculation}
\label{sec:exp}
Research into new areas often involves exploration 
predicated on intuitions 
that have yet to coalesce 
into crisp formal representations.
We recognize the role of speculation 
as a means for authors to impart intuitions 
that may not yet withstand the full weight 
of scientific scrutiny. 
However, 
papers often offer speculation in the guise of \emph{explanations}, 
which are then interpreted as authoritative 
due to the trappings of a scientific paper 
and the presumed expertise of the authors.

For instance, \citep{ioffe2015batch}
forms an intuitive theory around a concept 
called \emph{internal covariate shift}.
The exposition on internal covariate shift, 
starting from the abstract, 
appears to state technical facts.
However, key terms are not made crisp enough 
to conclusively assume a truth value.
For example, the paper states 
that batch normalization offers improvements 
by reducing changes in the distribution of hidden activations
over the course of training.
By which divergence measure is this change quantified?
The paper never clarifies, and some work suggests that this explanation of batch normalization may be off the mark \citep{santurkar2018does}.
Nevertheless, the speculative explanation given in \citep{ioffe2015batch} 
has been repeated as fact, 
e.g.~in \cite{noh2015learning}, which states, 
``It is well-known that a deep neural network is very hard
to optimize due to the internal-covariate-shift problem.''

We ourselves have been equally guilty 
of speculation disguised as explanation. 
In \citep{steinhardt2017certified}, 
JS writes that ``the high dimensionality and abundance of irrelevant features\ldots give the attacker more room 
to construct attacks'', 
without conducting any experiments 
to measure the effect of dimensionality on attackability. 
And in \citep{steinhardt2015reified}, 
JS introduces the intuitive notion of \emph{coverage} 
without defining it,
and uses it as a form of explanation, 
e.g.: ``Recall that one symptom of a lack of coverage 
is poor estimates of uncertainty 
and the inability to generate high precision predictions.'' 
Looking back, we desired
to communicate insufficiently fleshed out intuitions 
that were material to the work described in the paper,
and we were reticent to label a core part of our argument as speculative.

In contrast to the above examples, 
\citep{srivastava2014dropout} 
separates speculation from fact. 
While this paper, which introduced dropout regularization,
speculates at length on connections between dropout 
and sexual reproduction, 
a designated ``Motivation'' section 
clearly quarantines this discussion.
This practice avoids confusing readers
while allowing authors to express informal ideas. 

In another positive example, \citep{bengio2012practical} 
presents practical guidelines for training neural networks.
Here, the authors carefully convey uncertainty.
Instead of presenting the guidelines as authoritative, 
the paper states: 
``Although such recommendations come\ldots from years 
of experimentation and to some extent mathematical justification, 
they should be challenged. 
They constitute a good starting point\ldots 
but very often have not been formally validated, 
leaving open many questions 
that can be answered either by theoretical analysis 
or by solid comparative experimental work''.

\subsection{Failure to Identify the Sources of Empirical Gains}
\label{sec:contrib}

The machine learning peer review process 
places a premium on technical novelty. 
Perhaps to satisfy reviewers,
many papers emphasize both complex models (addressed here)
and fancy mathematics (see \S \ref{sec:mathiness}).
While complex models are sometimes justified, 
empirical advances often come about in other ways: 
through clever problem formulations, scientific experiments, 
optimization heuristics, data preprocessing techniques, 
extensive hyper-parameter tuning,
or by applying existing methods to interesting new tasks.
Sometimes a number of proposed techniques 
together achieve a significant empirical result.
In these cases, it serves the reader to elucidate 
which techniques are necessary 
to realize the reported gains. 

Too frequently, authors propose many tweaks 
absent proper ablation studies,
obscuring the source of empirical gains. 
Sometimes just one of the changes 
is actually responsible for the improved results. 
This can give the false impression 
that the authors did more work 
(by proposing several improvements),
when in fact they did not do enough 
(by not performing proper ablations).
Moreover, this practice misleads readers 
to believe that all of the proposed changes are necessary.

Recently, \citet{melis2018state} demonstrated 
that a series of published improvements,
originally attributed to complex innovations 
in network architectures, 
were actually due to better hyper-parameter tuning. 
On equal footing, vanilla LSTMs, hardly modified since 1997 \citep{hochreiter1997long}, 
topped the leaderboard. 
The community may have benefited more 
by learning the details of the hyper-parameter tuning
without the distractions.
Similar evaluation issues have been observed 
for deep reinforcement learning \cite{henderson2017deep} 
and generative adversarial networks \cite{lucic2017gans}. 
See \citep{sculley2018winner} 
for more discussion of lapses in empirical rigor and resulting consequences.

In contrast, many papers perform good ablation analyses \citep{lake2015human,liang2016neural,zagoruyko2016multipath,zhang2017tacred},
and even retrospective attempts 
to isolate the source of gains 
can lead to new discoveries 
\citep{chatfield2014return,santurkar2018does}. 
Furthermore, ablation is neither necessary nor sufficient 
for understanding a method, 
and can even be impractical given computational constraints. 
Understanding can also come from robustness checks 
(as in \citep{cotterell2018all}, 
which discovers that existing language models 
handle inflectional morphology poorly) 
as well as qualitative error analysis \citep{kwiatkowski2013scaling}. 

Empirical study aimed at understanding
can be illuminating even absent a new algorithm.
For instance, probing the behavior of neural networks led to identifying their susceptibility to adversarial perturbations
\cite{szegedy2013intriguing}. 
Careful study also often reveals limitations of challenge datasets while yielding stronger baselines.
\citep{chen2016thorough} studies
a task designed for reading comprehension 
of news passages and finds that 
$73\%$ of the questions can be answered 
by looking at a single sentence, 
while only $2\%$ require looking at multiple sentences 
(the remaining $25\%$ of examples were either ambiguous 
or contained coreference errors).
In addition,
simpler neural networks and linear classifiers
outperformed complicated neural architectures
that had previously been evaluated on this task. 
In the same spirit, \citep{zellers2018neural} analyzes and constructs a strong baseline for the Visual Genome Scene Graphs dataset.

\subsection{Mathiness}
\label{sec:mathiness}
When writing a paper early in PhD, 
we (ZL) received feedback from an experienced post-doc 
that the paper needed more equations. 
The post-doc wasn't endorsing the system, 
but rather communicating a sober view of how reviewing works.
More equations, even when difficult to decipher,
tend to convince reviewers of a paper's technical depth.

Mathematics is an essential tool for scientific communication, 
imparting precision and clarity 
when used correctly. 
However, not all ideas and claims 
are amenable to precise mathematical description, 
and natural language is an equally indispensible tool for communicating, 
especially about intuitive or empirical claims.

When mathematical and natural language statements are mixed without a clear accounting of their relationship, 
both the prose and the theory can suffer:
problems in the theory can be concealed by vague definitions,
while weak arguments in the prose 
can be bolstered by the appearance of technical depth.
We refer to this tangling of formal and informal claims as \emph{mathiness},
following economist Paul Romer who described 
the pattern thusly: ``Like mathematical theory, mathiness uses a mixture of words and symbols, but instead of making tight links, it leaves ample room for slippage 
between statements in natural language versus formal language'' \citep{romer2015mathiness}.

Mathiness manifests in several ways: 
First, some papers abuse mathematics 
to convey technical depth---to bulldoze rather than to clarify.
\emph{Spurious theorems} are common culprits, 
inserted into papers to lend authoritativeness 
to empirical results, 
even when the theorem's conclusions 
do not actually support the main claims of the paper. We (JS) are guilty of this in \citep{steinhardt2015learning}, 
where a discussion of ``staged strong Doeblin chains'' has limited relevance 
to the proposed learning algorithm, 
but might confer a sense of theoretical depth to readers.
 
The ubiquity of this issue is evidenced 
by the paper introducing the Adam optimizer  \citep{kingma2014adam}.
In the course of introducing an optimizer 
with strong empirical performance, 
it also offers a theorem regarding convergence in the convex case, 
which is perhaps unnecessary in an applied paper focusing on non-convex optimization. 
The proof was later shown to be incorrect 
in \citep{reddi2018convergence}.

A second issue is claims that are neither clearly formal nor clearly informal. For example, \citep{dauphin2014identifying} argues 
that the difficulty in optimizing neural networks stems not from local minima but from saddle points. 
As one piece of evidence, the work cites a statistical physics paper \citep{bray2007statistics} on Gaussian random fields
and states that in high dimensions ``all local minima [of Gaussian random fields] are likely to have 
an error very close to that of the global minimum'' (a similar statement appears in the related work of \citep{choromanska2015loss}).
This appears to be a formal claim, but absent a specific theorem it is difficult to verify the claimed result 
or to determine its precise content. Our understanding is that it is partially a numerical claim that the 
gap is small for typical settings of the problem parameters, as opposed to a claim that the gap vanishes 
in high dimensions. A formal statement would help clarify this. We note that the broader interesting point in \citep{dauphin2014identifying} 
that minima tend to have lower loss than saddle points is more clearly stated and empirically tested.

Finally, some papers invoke theory in overly broad ways, or make passing references to theorems 
with dubious pertinence. 
For instance, the \emph{no free lunch theorem} 
is commonly invoked as a justification 
for using heuristic methods without guarantees, 
even though the theorem does not formally preclude 
guaranteed learning procedures.

While the best remedy for mathiness is to avoid it, some papers go further with exemplary exposition. A recent paper \citep{bottou2013counterfactual} on counterfactual reasoning covers a large amount of mathematical ground in a down-to-earth manner, with numerous clear connections to applied empirical problems. This tutorial, written in clear service to the reader, has helped to spur work in the burgeoning community studying counterfactual reasoning for ML.

\subsection{Misuse of Language}
\label{sec:language}
We identify three common avenues of language misuse in machine learning: \emph{suggestive definitions}, \emph{overloaded terminology}, and \emph{suitcase words}.

\subsubsection{Suggestive Definitions}
In the first avenue, a new technical term is coined 
that has a suggestive colloquial meaning, 
thus sneaking in connotations without the need to argue for them. 
This often manifests in 
anthropomorphic characterizations of tasks (\emph{reading comprehension} \citep{hermann2015teaching} 
and \emph{music composition} \citep{mozer1994neural}) and techniques (\emph{curiosity} \citep{schmidhuber1991possibility} 
and \emph{fear} \citep{lipton2016combating}).
A number of papers name components of proposed models 
in a manner suggestive of human cognition, 
e.g.~``thought vectors'' \citep{kiros2015skip} 
and the ``consciousness prior'' \citep{bengio2017consciousness}.
Our goal is not to rid the academic literature of all such language; 
when properly qualified, these connections might communicate 
a fruitful source of inspiration. 
However, when a suggestive term is assigned technical meaning, each subsequent paper has no choice but to confuse its readers, 
either by embracing the term or by replacing it.

Describing empirical results with loose claims of ``human-level'' performance 
can also portray a false sense of current capabilities. 
Take, for example, the ``dermatologist-level classification of skin cancer''
reported in \cite{esteva2017dermatologist}.
The comparison to dermatologists conceals the fact that classifiers and dermatologists 
perform fundamentally different tasks. 
Real dermatologists encounter a wide variety of circumstances 
and must perform their jobs despite unpredictable changes.
The machine classifier, however, 
only achieves low error on i.i.d.~test data. 
In contrast, claims of human-level performance in \citep{he2015delving}
are better-qualified to refer to the ImageNet classification task
(rather than object recognition more broadly).
Even in this case, one careful paper (among many less careful \citep{esteva2017dermatologist, mnih2015human, taigman2014deepface})
was insufficient to put the public discourse back on track.
Popular articles continue to characterize modern image classifiers
as ``surpassing human abilities and effectively proving 
that bigger data leads to better decisions'' \citep{gershgorn2017quartz}, 
despite demonstrations that these networks rely on spurious correlations, 
e.g.~misclassifying ``Asians dressed in red'' as ping-pong balls \citep{stock2017convnets}.
 
Deep learning papers are not the sole offenders; misuse 
of language plagues many subfields of ML.
\citep{lipton2017fair} discusses 
how the recent literature on fairness in ML 
often overloads terminology borrowed from complex legal doctrine, 
such as \emph{disparate impact}, 
to name simple equations expressing 
particular notions of statistical parity.
This has resulted in a literature 
where ``fairness'', ``opportunity'', and ``discrimination'' 
denote simple statistics of predictive models, 
confusing researchers who become oblivious 
to the difference, 
and policymakers who become misinformed 
about the ease of incorporating ethical desiderata into ML.

\subsubsection{Overloading Technical Terminology}
A second avenue of misuse
consists of taking a term 
that holds precise technical meaning 
and using it in an imprecise or contradictory way.
Consider the case of \emph{deconvolution}, 
which formally describes the process of reversing a convolution, 
but is now used in the deep learning literature
to refer to transpose convolutions 
(also called up-convolutions) as commonly found in auto-encoders and generative adversarial networks.
This term first took root in deep learning 
in \citep{zeiler2010deconvolutional},
which does address deconvolution, 
but was later over-generalized to refer to
any neural architectures using upconvolutions  
\citep{zeiler2014visualizing,long2015fully}.
Such overloading of terminology can create lasting confusion.
New machine learning papers referring to deconvolution might be 
(i) invoking its original meaning, 
(ii) describing upconvolution, or 
(iii) attempting to resolve the confusion, 
as in \citep{hazirbas2017deep}, 
which awkwardly refers to ``upconvolution (deconvolution)''.

As another example, \emph{generative models} 
are traditionally models of either 
the input distribution $p(x)$ 
or the joint distribution $p(x,y)$. 
In contrast, discriminative models 
address the conditional distribution 
$p(y \mid x)$ 
of the label given the inputs. 
However, in recent works, ``generative model'' 
imprecisely refers to any model that produces realistic-looking structured data. 
On the surface, this may seem consistent 
with the $p(x)$ definition, 
but it obscures several shortcomings---for instance, 
the inability of GANs or VAEs to perform conditional inference 
(e.g.~sampling from $p(x_2 \mid x_1)$ 
where $x_1$ and $x_2$ are two distinct input features). 
Bending the term further, 
some discriminative models 
are now referred to as generative models 
on account of producing structured outputs \citep{yin2015neural},
a mistake that we (ZL) make in \citep{lipton2015generative}. 
Seeking to resolve the confusion 
and provide historical context, \citep{mohamed2016learning} distinguishes between \emph{prescribed} and \emph{implicit} generative models.

Revisiting batch normalization, 
\citep{ioffe2015batch} 
describes {covariate shift} 
as a change in the distribution of model inputs.
In fact, covariate shift 
refers to a specific type of shift 
where although the input distribution $p({x})$ 
might change, the labeling function $p(y|{x})$ 
does not \citep{gretton2009covariate}.
Moreover, due to the influence of \citep{ioffe2015batch}, 
Google Scholar lists batch normalization as the first reference on searches for ``covariate shift''. 

Among the consequences of mis-using language 
is that (as with generative models)
we might conceal lack of progress 
by redefining an unsolved task to refer to something easier. 
This often combines with suggestive definitions 
via anthropomorphic naming. \emph{Language understanding} 
and \emph{reading comprehension}, 
once grand challenges of AI, 
now refer to making accurate predictions on specific datasets \cite{hermann2015teaching}.

\subsubsection{Suitcase Words}
Finally, we discuss the overuse of \emph{suitcase words} in ML papers. 
Coined by Minsky in the 2007 book 
\emph{The Emotion Machine} \cite{minsky2007emotion},
suitcase words pack together a variety of meanings.
Minsky describes mental processes such as \emph{consciousness}, \emph{thinking}, \emph{attention}, \emph{emotion}, 
and \emph{feeling} that may not share ``a single cause or origin''.
Many terms in ML fall into this category.
For example, \cite{lipton2016mythos} notes that \emph{interpretability} 
holds no universally agreed-upon meaning,
and often references disjoint methods and desiderata. 
As a consequence, even papers that appear 
to be in dialogue with each other 
may have different concepts in mind. 

As another example, \emph{generalization} 
has both a specific technical meaning 
(generalizing from train to test) 
and a more colloquial meaning that is
closer to the notion 
of transfer (generalizing from one population to another) 
or of external validity (generalizing from an experimental setting to the real world) \citep{schram2005artificiality}. 
Conflating these notions leads to overestimating 
the capabilities of current systems.

Suggestive definitions and overloaded terminology 
can contribute to the creation of new suitcase words. 
In the fairness literature, where legal, philosophical, 
and statistical language are often overloaded, 
terms like \emph{bias} become suitcase words
that must be subsequently unpacked \citep{danks2017algorithmic}.

In common speech and as aspirational terms,
suitcase words can serve a useful purpose. 
Perhaps the suitcase word reflects an overarching concept that unites the various meanings. 
For example, \emph{artificial intelligence} 
might be well-suited as an aspirational name 
to organize an academic department. 
On the other hand, using suitcase words in technical arguments can lead to confusion. 
For example, \cite{bostrom2017superintelligence} 
writes an equation (Box 4) 
involving the terms \emph{intelligence} 
and \emph{optimization power}, 
implicitly assuming that these suitcase words 
can be quantified with a one-dimensional scalar.

\section{Speculation on Causes Behind the Trends}
\label{sec:causes}
Do the above patterns represent a trend, and if so, what are the underlying causes?
We \emph{speculate} that these patterns are on the rise 
and suspect several possible causal factors:
complacency in the face of progress, the rapid expansion of the community, 
the consequent thinness of the reviewer pool, 
and misaligned incentives of scholarship 
vs.~short-term measures of success.

\subsection{Complacency in the Face of Progress}

The apparent rapid progress in ML has at times engendered an
attitude that \emph{strong results excuse weak arguments}.
Authors with strong results may feel licensed to insert arbitrary
unsupported stories (see \S \ref{sec:exp})
regarding the factors driving the results,
to omit experiments aimed at disentangling those factors  (\S \ref{sec:contrib}), 
to adopt exaggerated terminology (\S \ref{sec:language}),
or to take less care to avoid
mathiness (\S \ref{sec:mathiness}).

At the same time, the single-round nature of the reviewing process may cause reviewers to feel they have no choice but 
to accept papers with strong quantitative findings.
Indeed, even if the paper is rejected, there is no guarantee the flaws
will be fixed or even noticed in the next cycle, 
so reviewers may conclude that accepting a flawed paper is the best option.

\subsection{Growing Pains}

Since around 2012, the ML community has expanded rapidly due to 
increased popularity stemming from 
the success of deep learning methods.
While we view the rapid expansion of the community 
as a positive development, 
it can also have side effects.

To protect junior authors, 
we have preferentially referenced our own papers 
and those of established researchers. 
However, newer researchers may be more susceptible 
to these patterns. 
For instance, authors unaware of previous terminology 
are more likely to mis-use or re-define language 
(\S \ref{sec:language}). 
On the other hand, experienced researchers fall into these patterns as well.

Rapid growth can also thin the reviewer pool, 
in two ways---by increasing the ratio of submitted papers to reviewers,
and by decreasing the fraction of experienced reviewers. 
Less experienced reviewers may be
more likely to demand architectural novelty, 
be fooled by spurious theorems, 
and let pass serious but
subtle issues like misuse of language, 
thus either incentivizing or enabling 
several of the trends described above. 
At the same time, experienced but over-burdened reviewers may revert to a ``check-list'' mentality,
rewarding more formulaic papers 
at the expense of more creative 
or intellectually ambitious work 
that might not fit a preconceived template. 
Moreover, overworked reviewers 
may not have enough time to fix---or even to notice---all of the issues in a submitted paper.

\subsection{Misaligned Incentives}

Reviewers are not alone in providing poor incentives for authors.
As ML research garners increased media attention
and ML startups become commonplace, 
to some degree incentives are provided 
by the press (``What will they write about?'') 
and by investors (``What will they invest in?'').
The media provides incentives for some of these trends.
Anthropomorphic descriptions of ML algorithms provide fodder for popular coverage.
Take for instance \citep{metz2014brain}, 
which characterizes an autoencoder as a ``simulated brain''.
Hints of human-level performance 
tend to be sensationalized in newspaper headlines, 
e.g. \cite{markoff2014human}, which describes a deep learning image captioning system as ``mimicking human levels of understanding''.
Investors too have shown a strong appetite 
for AI research, funding startups sometimes on the basis of a single paper. 
In our (ZL) experience working with investors,
they are sometimes attracted to startups 
whose research has received media coverage, a dynamic 
which attaches financial incentives to media attention. 
We note that recent interest in chatbot startups 
co-occurred with anthropomorphic descriptions 
of dialogue systems and reinforcement learners 
both in papers and in the media, 
although it may be difficult to determine 
whether the lapses in scholarship 
caused the interest of investors or vice versa.


\section{Suggestions}
\label{sec:suggestions}
Supposing we are to intervene to counter these trends, then how? 
Besides merely suggesting that each author abstain from these patterns, 
what can we do as a community 
to raise the level of experimental practice, exposition, and theory? 
And how can we more readily distill the knowledge of the community 
and disabuse researchers and the wider public 
of misconceptions?
Below we offer a number of preliminary suggestions 
based on our personal experiences and impressions. 

\subsection{Suggestions for Authors}

We encourage authors to ask ``what worked?''~and ``why?'', rather than just ``how well?''. 
Except in extraordinary cases \citep{krizhevsky2012imagenet}, 
raw headline numbers provide limited value 
for scientific progress 
absent insight into what drives them. 
Insight does not necessarily mean theory. 
Three practices that are common in 
the strongest empirical papers are 
\emph{error analysis},
\emph{ablation studies}, 
and \emph{robustness checks} 
(to e.g.~choice of hyper-parameters, 
as well as ideally to choice of dataset). 
These practices can be adopted by everyone 
and we advocate their wide-spread use. 
For some examplar papers, we refer the reader 
to the preceding discussion in \S \ref{sec:contrib}. \citep{langley1991experimental} also provides 
a more detailed survey of empirical best practices. 

Sound empirical inquiry need not be confined to tracing the sources of a particular algorithm's empirical gains; it can yield new insights 
even when no new algorithm is proposed.
Notable examples of this include 
a demonstration that neural networks trained by stochastic gradient descent 
can fit randomly-assigned labels \citep{zhang2016understanding}.
This paper questions the ability
of learning-theoretic notions of model complexity 
to explain why neural networks 
can generalize to unseen data. 
In another example, \citep{goodfellow2014qualitatively} 
explored the loss surfaces of deep networks,
revealing that straight-line paths in parameter space 
between initialized and learned parameters typically had
monotonically decreasing loss.
 
When writing, we recommend asking the following question: 
\emph{Would I rely on this explanation for making predictions or for getting a system to work?}
This can be a good test
of whether a theorem is being included to please reviewers 
or to convey actual insight. 
It also helps check whether concepts and explanations 
match our own internal mental model.
On mathematical writing, 
we point the reader to Knuth, Larrabee, and Roberts' 
excellent guidebook \citep{knuth1987mathematical}.

Finally, being clear about 
which problems are open and which are solved 
not only presents a clearer picture to readers, 
it encourages follow-up work 
and guards against researchers neglecting questions presumed (falsely) to be resolved.

\subsection{Suggestions for Publishers and Reviewers}

Reviewers can set better incentives by asking: \emph{``Might I have accepted this paper 
if the authors had done a worse job?''} 
For instance, a paper describing a simple idea that leads to improved performance, 
together with two negative results,
should be judged more favorably 
than a paper that combines three ideas together 
(without ablation studies)
yielding the same improvement.

Current literature moves fast at the expense of accepting flawed works for conference publication. One remedy could be to emphasize  authoritative retrospective surveys 
that strip out exaggerated claims and extraneous material, change anthropomorphic names to sober alternatives, standardize notation, etc. 
While venues such as \emph{Foundations and Trends in Machine Learning} 
already provide a track for such work, 
we feel that there are still not enough strong papers in this genre.

Additionally, we believe (noting our conflict of interest) 
that  critical writing ought to have a voice 
at machine learning conferences.
Typical ML conference papers 
choose an established problem 
(or propose a new one), 
demonstrate an algorithm and/or analysis, 
and report experimental results.
While many questions can be addressed in this way,
for addressing the validity of the problems 
or the methods of inquiry themselves, 
neither algorithms nor experiments 
are sufficient (or appropriate). 
We would not be alone in embracing greater critical discourse: 
in NLP, this year's COLING conference included 
a call for position papers ``to challenge conventional thinking'' \citep{coling2018}.

There are many lines of further discussion 
worth pursuing regarding peer review.
Are the problems we described mitigated or exacerbated by open review? 
How do reviewer point systems align with the values that we advocate?
These topics warrant their own papers and have indeed been discussed at length elsewhere \citep{langford_pr, lecun_pr, zoubin_pr}.

\section{Discussion}
\label{sec:discussion}
Folk wisdom might suggest not to intervene just as the field is heating up: 
\emph{You can't argue with success!} 
We counter these objections 
with the following arguments: 
First, many aspects of the current culture are \emph{consequences} 
of ML's recent success, not its \emph{causes}. In fact, many of the papers leading to the current success of deep learning were careful empirical investigations characterizing principles for training deep networks. This includes the advantage of random over sequential hyperparameter search \citep{bergstra2012random}, the behavior of different activation functions \citep{jarrett2009best,glorot2010understanding}, and an understanding of unsupervised pre-training \citep{erhan2010does}.

Second, flawed scholarship already negatively impacts the research community and broader public discourse. We saw in \S \ref{sec:troubling_trends} examples of unsupported claims being cited thousands of times, lineages of purported improvements being overturned by simple baselines, datasets that appear to test high-level semantic reasoning but actually test low-level syntactic fluency, and terminology confusion that muddles the academic dialogue. This final issue also affects the public discourse. For instance, 
the European parliament passed a report considering regulations to apply if ``robots become or are made self-aware'' \cite{eu2017}. While ML researchers are not responsible 
for all misrepresentations of our work, 
it seems likely that anthropomorphic language 
in authoritative peer-reviewed papers 
is at least partly to blame.

We believe that greater rigor in both exposition, science, and theory are essential for both scientific progress and fostering a productive discourse with the broader public. 
Moreover, as practitioners apply ML in critical domains such as health, law, and autonomous driving, a calibrated awareness of the abilities and limits of ML systems will enable us to deploy ML responsibly.
We conclude the paper by discussing 
several counterarguments and by providing historical context.

\subsection{Countervailing Considerations}
There are a number of countervailing considerations to the suggestions set forth above.
Several readers of earlier drafts of this paper noted
that \emph{stochastic gradient descent 
tends to converge faster than gradient descent}---in other words, perhaps a faster noisier process 
that ignores our guidelines for producing ``cleaner'' papers
results in a faster pace of research.
For example, the breakthrough paper 
on ImageNet  classification \citep{krizhevsky2012imagenet} 
proposes multiple techniques without ablation studies, 
several of which were subsequently determined 
to be unnecessary. 
However, at the time the results were so significant 
and the experiments so computationally expensive to run
that waiting for ablations to complete 
was perhaps not worth the cost to the community.

A related concern is that high standards might impede the publication of original ideas, which are more likely to be unusual and speculative. 
In other fields, such as economics,
high standards result in a publishing process 
that can take years for a single paper,
with lengthy revision cycles consuming resources 
that could be deployed towards new work.

Finally, perhaps there is value in \emph{specialization}:
the researchers generating new conceptual ideas 
or building new systems 
need not be the same ones who carefully collate and distill knowledge. 

We recognize the validity of these considerations, 
and also recognize that these standards are at times exacting.
However, in many cases they are straightforward to implement, 
requiring only a few extra days of experiments and more careful writing. 
Moreover, we present these as strong heuristics 
rather than unbreakable rules---if an idea cannot be shared 
without violating these heuristics, 
we prefer the idea be shared and the heuristics set aside.
Additionally, we have almost always found attempts 
to adhere to these standards to be well worth the effort. 
In short, we do not believe that the research community 
has achieved a Pareto optimal state on the growth-quality frontier. 

\subsection{Historical Antecedents}
The issues discussed here are neither unique to machine learning 
nor to this moment in time; 
they instead reflect 
issues that recur cyclically throughout academia. 
As far back as 1964, the physicist John R.~Platt  discussed related concerns in his paper on strong inference \citep{platt1964strong}, where he identified adherence to specific empirical standards as responsible for the rapid progress of molecular biology and high-energy physics relative to other areas of science.

There have also been similar discussions in AI. 
As noted in \S \ref{sec:introduction}, 
Drew McDermott \citep{mcdermott1976artificial} criticized 
a (mostly pre-ML) AI community in 1976 on a number of issues,
including suggestive definitions 
and a failure to separate out speculation from technical claims.
In 1988, Paul Cohen and Adele Howe \cite{cohen1988evaluation} 
addressed an AI community 
that at that point ``rarely publish[ed] performance evaluations'' 
of their proposed algorithms and instead only described the systems. 
They suggested establishing sensible metrics 
for quantifying progress, 
and also analyzing ``why does it work?'',
``under what circumstances won't it work?'' 
and ``have the design decisions been justified?'',
questions that continue to resonate today.
Finally, in 2009 Armstrong and co-authors \cite{armstrong2009improvements} 
discussed the empirical rigor 
of information retrieval research, 
noting a tendency of papers to compare 
against the same weak baselines, 
producing a long series of improvements 
that did not accumulate to meaningful gains.

In other fields, an unchecked decline in scholarship has led to crisis. 
A landmark study in 2015 suggested 
that a significant portion of findings 
in the psychology literature 
may not be reproducible \cite{open2015estimating}.
In a few historical cases, 
enthusiasm paired with undisciplined scholarship
led entire communities down blind alleys. 
For example, following the discovery of X-rays,
a related discipline on N-rays emerged \citep{nye1980n} 
before it was eventually debunked.

\subsection{Concluding Remarks}

The reader might rightly suggest 
that these problems are self-correcting.
We agree.
However, the community self-corrects 
precisely through recurring debate about 
what constitutes reasonable standards for scholarship.
We hope that this paper 
contributes constructively to the discussion.

\section{Acknowledgments}
\label{sec:acknowledgments}
We thank the many researchers, colleagues, and friends who generously shared feedback on this draft, 
including Asya Bergal, Kyunghyun Cho, Moustapha Cisse, Daniel Dewey, 
Danny Hernandez, Charles Elkan, Ian Goodfellow, 
Moritz Hardt, Tatsunori Hashimoto, Sergey Ioffe, Sham Kakade, David Kale, Holden Karnofsky, Pang Wei Koh, Lisha Li, 
Percy Liang, Julian McAuley, Robert Nishihara, 
Noah Smith, Balakrishnan ``Murali'' Narayanaswamy, 
Ali Rahimi, Christopher Ré, and Byron Wallace.
We also thank the ICML Debates organizers  
for the opportunity 
to work on this draft 
and for their patience 
throughout our revision process.

\bibliographystyle{plainnat}
\bibliography{main}
\end{document}